\documentclass[journal]{IEEEtran}
\ifCLASSOPTIONcompsoc
\else
\fi
\ifCLASSINFOpdf
\else
\fi

\usepackage{color}

\usepackage{graphicx}
\usepackage{subfigure}
\usepackage{algorithmic}
\usepackage[algoruled,vlined]{algorithm2e}
\usepackage{multirow}
\usepackage{footmisc}

\begin{document}
%
\title{Cross-Modal Learning via Pairwise Constraints}
%
%
%
%

\author{Ran~He, 
        Man Zhang, 
        Liang~Wang,~\IEEEmembership{Senior Member,~IEEE,}
        Ye Ji,~\IEEEmembership{Member,~IEEE,} and~ Qiyue Yin
\IEEEcompsocitemizethanks{\IEEEcompsocthanksitem Ran He, Man Zhang,
Liang Wang, and Qiyue Yin are with the Center for Research on Intelligent
Perception and Computing (CRIPAC) and National Laboratory of Pattern
Recognition (NLPR), Institute of Automation,
Chinese Academy of Sciences, Beijing, China 100190. 
E-mail: \{rhe, zhangman, wangliang, qyyin\}@nlpr.ia.ac.cn.
}
\IEEEcompsocitemizethanks{\IEEEcompsocthanksitem Ye Ji is with the department
of Control Science and Engineering, Shandong University, Jinan, China 250100. 
E-mail: jeeye@163.com.
}\thanks{}}

%
%

\markboth{Journal of \LaTeX\ Class Files,~Vol.~6, No.~1, January~2007}%
{Shell \MakeLowercase{\textit{et al.}}: Bare Demo of IEEEtran.cls for Computer Society Journals}
%


\IEEEcompsoctitleabstractindextext{%
\begin{abstract}
   In multimedia applications, the text and image components in a
   web document form a pairwise constraint that potentially indicates the same
   semantic concept. This paper studies cross-modal
   learning via the pairwise constraint, and aims to find the
   common structure hidden in different
   modalities. We first propose a compound regularization framework
   to deal with the pairwise constraint, which can be used as a general
   platform for developing cross-modal algorithms. For unsupervised
   learning, we propose a cross-modal subspace clustering method to learn a
   common structure for different modalities.
   For supervised learning, to reduce the semantic gap and the outliers
   in pairwise constraints, we propose a cross-modal
   matching method based on compound $\ell_{21}$ regularization along
   with an iteratively reweighted algorithm to find the global optimum.
   Extensive experiments demonstrate the benefits of joint text and image
   modeling with semantically induced pairwise constraints, and show that the proposed
   cross-modal methods can further reduce the semantic gap between
   different modalities and improve the clustering/retrieval accuracy.
\end{abstract}

}

\maketitle

\IEEEdisplaynotcompsoctitleabstractindextext

%
\IEEEpeerreviewmaketitle

\section{Introduction}
\IEEEPARstart{M}{ultimedia} information of data presents
diversified combination of different forms, such as text, video,
still images, and live TV. It is intrinsically multi modal
\cite{Bekkerman:2007} and often requires a web document corpus with
paired text and images (or other forms). One
of its fundamental tasks is to learn cross-modal information from
multiple content modalities. Recently, the terms 'cross-modal'\cite{Rasiwasia:2010}\cite{YJia:2011},
'multi-modal'\cite{Bekkerman:2007}\cite{HTong:2005} and 'multi-view'\cite{MKan:2012}\cite{YLuo:2013} are all used for multimedia information processing, and the word 'modality' has different interpretation in different applications. In this paper, multiple modalities (e.g., text and images) are assumed to have a loose relation, and each modality, which gives a different aspect of  multimedia information, has a dependent relationship to other modalities~\cite{Bekkerman:2007}.

\begin{figure*}[t]
\begin{center}
    \subfigure[]{\includegraphics[width=0.46\linewidth]{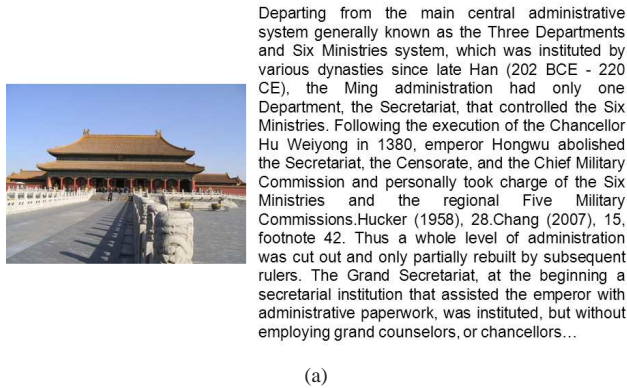}}
   \subfigure[]{\includegraphics[width=0.46\linewidth]{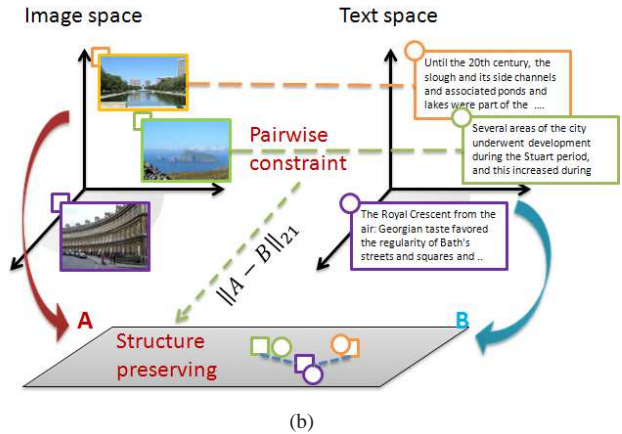}}
\end{center}
   \caption{Cross-modal learning via pairwise constraint. (a) An example of image-text pair
   from the Wiki dataset~\cite{Rasiwasia:2010}. The image has a loosely related text description
   \cite{YJia:2011}. (b) The pairwise constraint induces two
   basic problems: how to learn a common structure for two modalities? and how to preserve
   the structure during learning? Data sets A and B are the
   representations of two modalities respectively in a hidden space, in which
   paired points are close meanwhile the pairwise constraint induced
   structure between data points are preserved. $\ell_{21}$-norm is adpoted
   to improve robustness and reduce the semantic gap between low-level
   features and high-level concepts.}
\label{fig:frame}
\end{figure*}

In multimedia information processing, one popular strategy is to apply paired samples from different modalities to learn a common latent structure (or space) and then to perform clustering or retrieval. The paired samples refer to the samples from different modalities that belong to the same semantic unit( e.g., text and image in a web document, image features and associated tags for an image) and form a pairwise constraint for different modalities as shown in Fig.~\ref{fig:frame}. This pairwise constraint problem has also drawn much attention in other applications such as multi-pose face recognition \cite{MKan:2012}\cite{DLin:2006}, biometric verification \cite{ZCui:2012}, multilingual retrieval~\cite{Yogatama:2009}, semi-supervised learning \cite{SSun:2010} and dictionary learning \cite{HGuo:2012}\cite{KJia:2013}.

Although different methods have been developed for cross-modal
learning and pairwise constraints have drawn
much attention in different communities, to the best of our
knowledge, there may be not any existing works to fully explore
pairwise constraint problems from a unified viewpoint. In multimedia
information processing, since different modalities are in one single
document (e.g., the text and image in a web document as shown in
Fig.~\ref{fig:frame}), they form a pairwise constraint and give
different aspects of multimedia information. However, as shown in
Fig. \ref{fig:frame} (a), the paired samples often have a loose
relationship \cite{YJia:2011}, which makes the cross-modal learning
with pairwise constraints more challenging. In addition, since there
is a semantic gap between low-level features and high-level concepts,
the pairwise constraint often leads to two basic problems for
learning methods as shown in Fig. \ref{fig:frame} (b):
%

1) How to learn a common structure for two modalities? Because of
the semantic gap, one can obtain different structures from different
modalities. For example, points $x_i$ and $x_j$ may be neighbors in
the text space whereas they are not in the image space. However,
these structures must be the same neighborhood structure due to the
pairwise constraint.

2) How to preserve the structure during learning? For a cross-modal
learning algorithm, two modalities in the learned subspace (or
representation) should satisfy the original pairwise constraint and
preserve the original neighborhood structure.

This paper systematically studies the pairwise constraint and its
induced structure preserving problems, and accordingly proposes a
general regularization framework to find the common structure hidden
in different modalities~\footnote{Here we focus
only on the pairwise constraint of the multiple modalities from web
documents. Different modalities indicate the same semantic concept and have the same
nearest neighborhood structure.}. In particular, for unsupervised
learning, we propose a cross-modal subspace learning method, in
which different modalities share a common structure. A simple but
efficient algorithm is further developed to solve the subspace
clustering problem. For supervised learning, to reduce the semantic
gap and the outliers in pairwise constraints, we propose a cross-modal matching method based on
compound $\ell_{21}$ regularization, which can be efficiently solved
by an iteratively reweighted algorithm. At each iteration, the
compound $\ell_{21}$ regularization problem is simplified to a least
squares problem. Extensive
experiments on several widely used databases
demonstrate the benefits of joint text and image modeling with
pairwise constraints.

The main contribution of this work lies in three-fold:

1) We make a systematic investigation on the pairwise constraint in
cross-modal learning, and design a general regularization
framework, which can be used as a general platform for developing
unsupervised and supervised learning algorithms.

2) For unsupervised learning, to the best of our knowledge, it is
the first time to extend the linear presentation based subspace
clustering
methods~\cite{Elhamifar:2013}\cite{GLiu:2010}\cite{GLiu:2011}\cite{CLiu:2012}
to deal with the multi-modal case. Experimental results show that
the clustering accuracy can be improved when multi-modal
information is used.

3) For supervised learning, a $\ell_{21}$ regularization method is
proposed for cross-modal matching, which can handle intra-class
variation, pairwise constraint and structure preserving at the same
time. It obtains the state-of-the-art results in the Wiki
text-image dataset \cite{Rasiwasia:2010}.

The rest of this paper is organized as follows. In
Section~\ref{sec:pre}, we briefly review existing cross-modal (or
multi-modal) learning methods. In Section~\ref{sec:model},
we propose a general framework for cross-modal learning. In
particular, a cross-modal subspace clustering method and a
cross-modal matching method are developed
for unsupervised and supervised learning respectively. 
Section \ref{sec:exp} provides a series of experiments to
systematically evaluate the effectiveness of the proposed methods,
prior to the summary of this paper in Section~\ref{sec:con}.

\section{Related work \label{sec:pre}}
Our proposed unsupervised and supervised multi-modal learning methods
correspond to clustering and retrieval tasks respectively.
In this section, we accordingly review some related cross-modal methods in clustering and matching tasks.

\subsection{Clustering on multi-modal data}
For clustering, Basu et al. \cite{Basu:2004} presented a
pairwise constrained clustering framework together with a method to
select informative pairwise constraints to improve clustering
performance. Cho et al. \cite{HCho:2004} proposed a minimum
sum-squared residue co-clustering method for gene expression data.
Tong et al. \cite{HTong:2005} studied a graph based multi-modality
clustering algorithm to group multiple modalities. Based on
combinatorial Markov random field, Bekkerman and Jeon
\cite{Bekkerman:2007} developed a multi-modal clustering method for
multimedia collections. Based on spectral clustering, Yogatama and
Tanaka-Ishii~\cite{Yogatama:2009} presented a multilingual spectral
clustering to merge two language spaces via pairwise constraints;
and by combining co-regularization for multiple
views~\cite{Vindhwani:2005}\cite{SSun:2010}, Kumar et al.
\cite{Kumar:2011}\cite{Kumar:2011_2} presented co-training and
co-regularized approaches for multi-view spectral clustering
respectively. In addition, deep networks were used in
\cite{YKang:2012}\cite{Srivastava:2012} to learn shared
representations for multi-modal data. Wang et al. \cite{BWang:2012}
resorted to the cross diffusion process to fuse multiple metrics. The
authors in \cite{YYang:2012}\cite{Elhamifar:2012} apply the
similarity (or dissimilarity) measures w.r.t. pairwise constraints
to exemplar clustering and exemplar finding tasks respectively.
Base on structured sparsity, Wang et al. \cite{HWang:2013} proposed
to learn cluster indication matrix and then used K-means to perform
clustering. Hua and Pei~\cite{MHua:2012} proposed bottom-up and
top-down methods for mutual subspace clustering.

Recently, structure prior information (such as sparse~\cite{Elhamifar:2013}, low-rank~\cite{GLiu:2010}, or collaborative~\cite{CLiu:2012})
has shown to be effective for single-modality clustering and often results in
better clustering accuracy, which drives us to develop new multi-modal
clustering methods based on the structure prior information.

\subsection{Cross-modal matching}
For retrieval, the most famous cross-modal methods to
obtain a common space for multiple modalities are canonical
correlation analysis (CCA)~\cite{Kim:2007}\cite{Rasiwasia:2010} and
partial least squares (PLS)~\cite{Sharma:2011}\cite{YChen:2012},
which learn transformations to project each modality into a common space.
Since CCA does not use label information of multiple modalities,
multi-view discriminant analysis \cite{Diethe:2008}\cite{MKan:2012}
are further developed to make use of label information. By means of
locality preserving, Sun and Chen \cite{TSun:2007} presented
locality preserving CCA, and Quadrianto and Lampert
\cite{Quadrianto:2011} developed multi-view neighborhood preserving
projections. Based on bilinear models \cite{Tenenbaum:2000},
Sharma et al. \cite{Sharma:2012} further extended multi-view
discriminant analysis to a generalized multi-view analysis in terms
of graph embedding \cite{Yan:2007}.  Weston et al.
\cite{Weston:2010}\cite{Lucchi:2012} tried to learn common
representation spaces for images and their annotations. Sang and Xu
\cite{JSange:2012} provided a new perspective of multi-modal video
analysis by exploring the pairwise visual cues for constrained topic
modeling. Based on structured sparsity, Zhuang et al. \cite{YZhuang:2013}
proposed a supervised coupled dictionary learning method for multi-modal retrieval.

In other multi-modal applications, Lin and Tang
\cite{DLin:2005}\cite{DLin:2006} resorted to subspace learning for
inter-modality face recognition. Ye et al. \cite{GYe:2012} applied
pairwise relationship matrix for robust late fusion. Cui et al.
\cite{ZCui:2012} developed a pairwise constrained multiple metric
learning method for face verification. In dictionary learning,
the authors in \cite{HGuo:2012}\cite{KJia:2013}\cite{YZhuang:2013} resorted to paired
samples to learn discriminative dictionaries for image
classification. Kulis et al. \cite{Kulis:2012} proposed asymmetric
kernel transforms for cross domain adaptation. Chen et al.~\cite{XChen:2012}
proposed a general framework to deal with semi-paired and semi-supervised multi-view data, which
combines both structural information and discriminative information. By considering side information,
Qian et al.~\cite{QQian:2013} proposed a multi-view classification method with cross-view must-link
and cannot-link constraints. Xu et al. \cite{CXu:2014} extended the
theory of the information bottleneck to learn from examples
represented by multi-view features. Jiang et al. \cite{YJiang:2013} developed a novel
semi-supervised unified latent factor learning method for partially
labeled multi-view data.

Although many learning algorithms have been developed for cross-modal
problems and pairwise constraints have been studied in cross-modal
learning, there is still not any systematic work to fully explore pairwise
constraints. Hence a general cross-modal learning framework may be potentially
useful for future research.

\section{Cross modal learning via pairwise constraints\label{sec:model}}
In this section, we study unsupervised and supervised methods for
cross-modal learning via the pairwise constraint. Although the
proposed ideas can also be used for multi-view learning
\cite{DLin:2006}\cite{Sharma:2012}, multi-task learning
\cite{JHe:2012}\cite{JZhou:2012} and other combinations of content
modalities \cite{HCho:2004}\cite{JSange:2012}, here we restrict our
study to documents containing images and text as in
\cite{Rasiwasia:2010}\cite{YChen:2012}. The goal is to utilize the
pairwise constraint to improve learning results.

\subsection{A general framework}
Web documents often pair a body of text with a number of images
\cite{Rasiwasia:2010}, which form pairwise constraints for
cross-modal learning. For simplicity, we only discuss
the case that one document contains only one image and a body of
text such that there is only one pairwise constraint in one
document. Let $X_A \in R^{d_A \times n}$ and $X_B \in R^{d_B \times
n}$ are two modalities of documents that contain components of
images and text respectively, $n$ is the number of documents, and
$d_A$ and $d_B$ are feature dimensions of images and text
respectively. We expect to learn subspaces $U_I$ ($I\in\{A,B\}$) and
their corresponding embedding $Z_I$ such that $U_IZ_I$ can mostly
agree with $X_I$. In addition, due to the semantic gap between different modalities,
the representation abilities of multi-modal features are imbalanced so that a unique $Z_I=Z$ cannot represent different modalities well. To alleviate this problem, we also expect embedding $Z_A$ and $Z_B$ are closer as much as possible according to the pairwise constraint.
Hence, we have the following cross-modal learning problem in general,
\begin{eqnarray} \label{eq:prob}
\min_{U_I,Z_I}\sum\limits_{I } (w_I\left\| {X_I  - U_IZ_I}
\right\|_F^2 +\lambda_1 f_s(U_I,Z_I)) \\
+ \lambda_2\left\|
{Z_A  - Z_B} \right\|, \nonumber
\end{eqnarray}
where $w_I$, $\lambda_1$ and $\lambda_2$ are constants, $\left\| .
\right\|_F$ is matrix Frobenius norm, $f_s(.)$ is a function about
$U_I$ and $Z_I$ to preserve the structure, and $\left\| . \right\|$ is a
potential norm to handle the property of $Z_A-Z_B$, e.g.,
$\ell_1$-norm, $\ell_{21}$-norm or nuclear norm. The minimization problem
in (\ref{eq:prob}) is non-convex w.r.t $\{U_I,Z_I\}$. When $U_I$ or $Z_I$ is fixed, (\ref{eq:prob}) becomes the co-regularized least squares regression problem \cite{Sindhwani:2005}\cite{Brefeld:2006}\cite{Sindhwani:2008}; and if $\ell_1$-norm is used in $\left\|{Z_A  - Z_B} \right\|$, (\ref{eq:prob}) can be viewed as an extension of the pairwise lasso problem~\cite{SSun:2010}\cite{Petry:2011}.

The first item in (\ref{eq:prob}) is a data adaptation item, the
second item controls the complexity of subspace $U_I$, and the last
one models the pairwise constraint between two modalities. Both
the second and third items facilitate structure preserving. For
example, $f_s(.)$ and $\left\| . \right\|$ on ${U_I}$ can be nuclear
norm, structured sparsity induced norm, or a graph Laplacian
regularization in Subsection~\ref{sec:sup}. The second item aims to
preserve the structure of each modality and the third item ensures
that the structures of different modalities are similar. For unsupervised learning,
$U_I$ can be a dictionary to express modality $X_I$, and $Z_I$ can
be graph affinity matrix for each modality
\cite{HTong:2005}\cite{Kumar:2011}\cite{Kumar:2011_2}; and for
supervised learning, $U_I$ can be a discriminative projection matrix
to project different modalities into a common subspace for
cross-modal retrieval/classification, and $Z_I$ can be a group
indicator matrix to represent different semantic
groups~\cite{Sharma:2012}\cite{MKan:2012}. In addition,
(\ref{eq:prob}) can be viewed as a dictionary learning problem
with paired samples in \cite{HGuo:2012}\cite{KJia:2013}, in which
$U_I$ and $Z_I$ are a dictionary and a coefficient matrix
respectively. In the following two subsections, we will detail the
proposed model in (\ref{eq:prob}) for unsupervised learning and
supervised learning respectively.


\subsection{Unsupervised learning}
Clustering is one of main components in multimedia management
systems. For multimedia information, an effective clustering system
aims to handle complex structures and discover common
representations of multimedia documents \cite{Bekkerman:2007}.
Here, we focus on the problem of bridging multi-modal spaces for web document clustering. We are given web documents with different modalities (e.g., text and image) and asked to group them into clusters so that web documents from the same topic are grouped together.

Inspired by the recent advances in subspace clustering (or
segmentation) \cite{Elhamifar:2013}, we consider a diagonal
constraint $diag(Z_I)=0$ and set subspace $U_I$ to be $X_I$. In addition,
we expect that $Z_I$ can reflect some data structures, such as sparse and low-rank.
Hence we let $f_s(U_I,Z_I)=\left\| {Z_I} \right\|$ that is a
structure preserving item to make $Z_I$ be collaborative, sparse or low-rank as in subspace
clustering \cite{Elhamifar:2013}\cite{GLiu:2010}.
Then (\ref{eq:prob}) takes the following form,
\begin{equation} \label{eq:ucp}
\min _{Z_I} \sum\limits_{I } {( w_I \left\| {X_I  - X_I Z_I }
\right\|_F^2 +\lambda_1 \left\| {Z_I} \right\|  ) }+ \lambda _2 \left\| {Z_A  - Z_B } \right\|,
\end{equation}
where $Z_I$ indicates a graph affinity matrix for each modality as in \cite{HTong:2005}\cite{Kumar:2011}\cite{Kumar:2011_2}
(That is, one modality is represented as one independent graph \cite{HTong:2005}). The last item
$\left\| {Z_A^T  - Z_B^T } \right\|$ makes each graph $Z_I$ to agree with each other under
the constraint formulated by a norm $\left\|. \right\|$, such as $\ell_1$, $\ell_{2}$ and nuclear norms.
Here we consider a simple case of $\ell_2$-norm. Then we have,
\begin{equation} \label{eq:ucp1}
\min _{Z_I} \sum\limits_{I } {( w_I \left\| {X_I  - X_I Z_I }
\right\|_F^2 +\lambda_1 \left\| {Z_I} \right\|_F^2  ) }+ \lambda _2 \left\| {Z_A  - Z_B } \right\|_F^2.
\end{equation}
Furthermore, let $Z=(Z_A+Z_B)/2$, we can derive that,
$$Z_A - Z_B=2Z_A-(Z_B+Z_A)=2Z_A-2Z.$$
Hence (\ref{eq:ucp1}) can be reformulated as,
\begin{equation} \label{eq:ucp2}
\min _{Z_I, Z} \sum\limits_{I } {( w_I \left\| {X_I  - X_I Z_I }
\right\|_F^2 +\lambda_1 \left\| {Z_I} \right\|_F^2  + \lambda _3 \left\| {Z_I  - Z } \right\|_F^2) },
\end{equation}
where $\lambda_3=2\lambda_2$.

If we set the derivative of (\ref{eq:ucp2}) with respect to $Z$ equal to zeros,
we obtain that the optimal solution of $Z$ takes the form,
\begin{equation} \label{eq:umean}
  Z^* = (Z_A^* +Z_B^*)/2,
\end{equation}
where $Z_A^*$ and $Z_B^*$ are optimal solutions of $Z_A$ and $Z_B$ respectively.

The optimal solution of (\ref{eq:ucp2}) can be obtained in an alternating minimization way.
We can set the derivative of (\ref{eq:ucp2}) with respect to $Z_I$ equal to zeros respectively
and find a solution of $Z_I$. Considering that the diagonal
constraint $diag(Z_I)=0$ in subspace clustering, we can compute $Z_I$ as follows,
\begin{equation} \label{eq:uzi}
z_i^I  = (w_I \bar X_I^T \bar X_I  + \lambda _1 I + \lambda _3 I)^{ - 1} (w_I \bar X_I^T x_i^I  + \lambda _3 z_i^* ),
\end{equation}
where $z_i^I \in {R^{\left( {n - 1} \right) \times 1}}$ is the $i$-th column of $Z_I$ excluding $Z_{I(ii)}=0$,
$\bar X_I \in {R^{{d_I}\times\left( {n - 1} \right)}}$ indicates all data points in $X_I$ excluding $x_i^I$,
$x_i^I$ indicates the $i$-th data point in $X_I$, and $z_i^*$ is the $i$-th column of $Z$ excluding $Z_{ii}$.

\begin{algorithm}[]
\caption{Cross-modal Subspace Clustering (CSC) with Pairwise Constraint \label{Alg:CSC}}
\KwIn{$X_A \in R^{d_A \times n}$, $X_B \in R^{d_B \times n}$, parameter $\lambda _1$ and $\lambda _3$, the number of clusters $c$}
\KwOut{Groups of the dataset}
\begin{algorithmic}[1]
\STATE $\;$ $\;$ For $i=1:n$
\STATE $\;$ $\;$ $\;$ $\;$ Repeat:
\STATE $\;$ $\;$ $\;$ $\;$ $\;$  Solve $z_i^I$ according to (\ref{eq:uzi});
\STATE $\;$ $\;$ $\;$ $\;$ $\;$  Solve $Z$ according to (\ref{eq:umean});
\STATE $\;$ $\;$ $\;$ $\;$ Until convergence
\STATE $\;$ $\;$ End
\STATE $\;$ $\;$ Define the affinity matrix $A = \frac{1}{2}({\left| {{Z^T}} \right|} + \left| {{Z}} \right|)$.
\STATE Apply the Ncuts \cite{Ncuts} to the affinity matrix $A$.
\end{algorithmic}
\end{algorithm}

Algorithm~\ref{Alg:CSC} summarizes the procedure of our cross-modal subspace clustering method. Since we can compute $Z$ by computing each $z^*_i$ independently, the computation of $Z$ can be separated and paralleled. To further reduce computational costs of each $z^*_i$, we can minimize (\ref{eq:uzi}) only from $x_i^I$'s nearest neighborhood samples. As a result, the computation mainly depends on the iteration in Algorithm~\ref{Alg:CSC} rather than the number of data. When the number of data tends to be large, the major computational cost of Algorithm~\ref{Alg:CSC} depends on its clustering step.

\subsection{Supervised learning \label{sec:sup}}
In multimedia retrieval applications, a practical cross-modal
retrieval problem often includes two tasks \cite{Rasiwasia:2010}:
one is to retrieve images in response to a query text; and the other
is to retrieve text documents in response to a query image.
Recently, some learning methods are developed to learn common
representations \cite{Srivastava:2012}\cite{YKang:2012} or
discriminative subspaces
\cite{YChen:2012}\cite{Sharma:2012}\cite{MKan:2012} for cross-modal
problems. Inspired by these methods, we aim to learn two subspaces
$U_A$ and $U_B$ in which the projected data are most discriminative
and relevant. Furthermore, we resort to the indicator matrix (or spectral matrix) $Y
\in R^{n \times c}$ in linear discriminant analysis (LDA)
\cite{Cai:2007}\cite{Ye:2007} as the hidden space $Z$ in
(\ref{eq:prob}) for two modalities. Then we have the following loss
function,
\begin{equation} \label{eq:sfu}
f(U_A ,U_B ) \buildrel\textstyle.\over= \sum\nolimits_I {\left\|
{U_I^T X_I  - Y} \right\|} _F^2,
\end{equation}
where
\begin{equation} \label{eq:sy}
Y_{il}  = \left\{ {\begin{array}{*{20}c}
   1 & {{\rm{if \;}}x_i {\rm{\; belongs \; to \; the \; }}l-{\rm{th \; class}}}  \\
   0 & {{\rm{otherwise}}}  \\
\end{array}} \right. ,
\end{equation}
and $l\in \{1, \dots, c\}$. The definition of (\ref{eq:sy}) is the same as
(15) in \cite{Cai:2007} and (12) in \cite{Ye:2007}. Note that, for a specific application, the label matrix
$Y$ can be any other spectral matrices of graph embedding methods
\cite{Cai:2007}. If we only consider one modality in (\ref{eq:sfu}) and $Y$ is the indicator matrix of LDA,
(\ref{eq:sfu}) becomes the least square formulation of LDA in terms
of graph embedding \cite{Cai:2007}, which makes use of within-class
and between-class variations for a discriminative purpose. Hence,
(\ref{eq:sfu}) can also be viewed as a natural extension from single
modality LDA to multiple modalities.

\begin{figure}[t]
\begin{center}
\includegraphics[width=80mm]{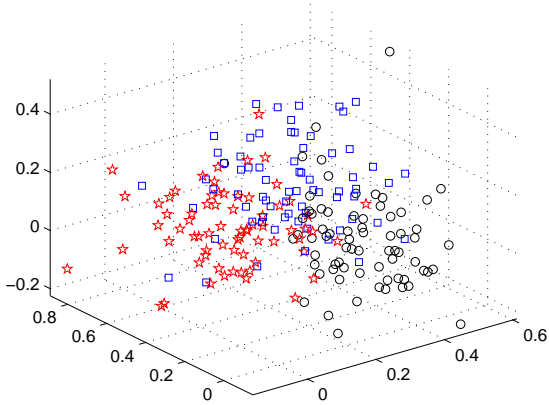}
\end{center}
   \caption{A three-dimensional representation of the set of Wiki image data using (\ref{eq:sfu}).
   For the first three categories, 70 image samples per category are randomly selected from the
   Wiki training dataset. Then (\ref{eq:sfu}) is applied to obtain the lower dimensional representation $U_BX_B$.
   \label{fig:wiki}}
\end{figure}


In (\ref{eq:sfu}), if each $U_I^T X_I$ is close enough to $Y$,
$U_A^T X_A$ and $U_B^T X_B$ will be close to each other so that cross-modal retrieval on $X_A$
and $X_B$ will be very accurate. However, because of semantic gap between high-level semantic concept and low-level
features, $U_A^T X^A$ and $U_B^T X^B$ may be not close to $Y$. In real-world cross-modal retrieval tasks,
it is almost impossible to find two subspaces $U_A$ and $U_B$ so that $U_A^T X_A = U_B^T X_B=Y$. Particularly,
a pair of data $U_A x_i^A$ and $U_B x_i^B$ may be far away from each other. Because (\ref{eq:sfu}) is a least
square formation of LDA, it only facilitates clustering the data from the same class as well as making the data from different
classes be far away. Fig.~\ref{fig:wiki} gives an illustration on the Wiki text-image dataset. We observe that
although the images from the first three categories tend to be clustered in the three dimensional subspace, each
projection point $U_Bx_i^B$ is not close to the indicator matrix $Y$.

Considering that structure preserving is often useful in graph embedding methods, we substitute $f_s(U_I,Z_I)$
in (\ref{eq:prob}) with a structure preserving item, i.e., $f_s(U_I,Z_I)=\sum\nolimits_I
{ \sum\nolimits_{\{i,j\}} {w_{ij}\left\| {U_I^T(x_i^I  - x_j^I) } \right\|_2 } }$. $w_{ij}$ is a constant that
indicates the relationship between $x^I_i$ and $x^I_j$. Because of the semantic gap, the
observed relationship between $x^A_i$ and $x^A_j$ may be different
from that between $x^B_i$ and $x^B_j$. A simple way to solve this problem is to concatenate the
feature vectors in each modality, and then resort to a weighting calculation strategy in graph embedding
to learn a high-level relationship.

Although $f(U_A,U_B)$ in (\ref{eq:sfu}) clusters data and $f_s(U_I,Z_I)$ preserves data structure, both
$f(U_A,U_B)$ and $f_s(U_I,Z_I)$ can not ensure each pair of modal data be close to each other in the
projection subspace. Hence we need to introduce an item to make each pair of modal data follow pairwise
constraints. By combining all things together, we obtain the following regularization
problem via (\ref{eq:prob}),
\begin{eqnarray}\label{eq:sobj}
\mathop {\min }\limits_{U_A,U_B} f(U_A,U_B)+\lambda _1\sum\limits_I
{ \sum\limits_{i,j} {w_{ij}
\left\| {U_I^T(x_i^I  - x_j^I) } \right\|_2 } } \\
+ \lambda _2 \left\| {X_A^T U_A  - X_B^T U_B }
\right\|_{21},\nonumber
\end{eqnarray}
where $\left\|M\right\|_{21}\buildrel\textstyle.\over=\sum\nolimits_i
{\sqrt {\sum\nolimits_j {{m}_{ij}^2 } } }$ is the $\ell_{21}$
norm of matrix $M$~\cite{RHe:2012}, and $w_I$ in (\ref{eq:prob}) is equal to 1.

The multi-modal problem in (\ref{eq:sobj}) can be viewed as an extension and variant of the co-regularized least squares regression \cite{Sindhwani:2005}\cite{Brefeld:2006}\cite{Sindhwani:2008}. The second term in
(\ref{eq:sobj}) can be viewed as a weighted $\ell_{21}$-norm and
used to preserve the structure of original data. Because of the
semantic gap between low-level feature and high-level semantic
concept, $\ell_{21}$-norm is used to make the objective function
focus on some important relationships between $x^I_i$ and $x^I_j$.
In addition, a $\ell_{21}$-norm is imposed on the third term in
(\ref{eq:sobj}) such that pairwise constraint is preserved in
learned subspaces meanwhile the outliers from inaccurate or
corrupted pairs are removed. This $\ell_{21}$-norm can be viewed as an extension of the $\ell_1$-norm in sparse multi-view co-regularized least squares \cite{SSun:2010}. Note that the outliers in pairwise constraints widely exist in text-image retrieval applications. Since the representation abilities of text and image features are imbalanced, it is difficult to find two maps $U_A$ and $U_B$ to make each pair $X_A^T U_A$ and $X_B^T U_B$ closer. The low accuracy of cross-modal retrieval in Section~\ref{sec:exps} also demonstrates that most of paired data are not well matched.


It is difficult to directly minimize the compound $\ell_{21}$
objective function in (\ref{eq:sobj}) because $\ell_{21}$-norm is not continuous on the origin.
Fortunately, the iteratively reweighted method \cite{FNie:2010}, the conjugate function method \cite{SSun:2010},
and the half-quadratic minimization method \cite{RHe:2012} have been developed to solve $\ell_{2,1}$-norm minimization problems.
According to \cite{FNie:2010}\cite{RHe:2012}, the augmented objective function of
(\ref{eq:sobj}) takes the form,
\begin{eqnarray}
J(U_I ,p^I ,q) \buildrel\textstyle.\over= f(U_A ,U_B ) + \lambda _1
\sum\limits_{I,i,j} {w_{ij}p^I_{ij} \left\| {U_I^T (x_i^I
-x_j^I )} \right\|_2^2} \nonumber\\
+\lambda_2 tr((U_A^T X_A-U_B^T X_B )Q(X_A^TU_A - X_B^TU_B))
\nonumber
\end{eqnarray}
where $tr(.)$ is matrix trace operator, $p_{ij}^I$ and $q_i$ are
auxiliary variables that depend on $U_A$ and $U_B$, and $Q$ is the diagonal matrix whose
$i$-th diagonal element is $q_i$. According to half-quadratic
minimization \cite{RHe:2013}\cite{RHe:2012}, one can
minimize the augmented objective function as follows,
\begin{eqnarray}
p^{It}_{ij}=1/{\left\| {(U_I^{t-1})^T (x_i^I
-x_j^I )} \right\|_2}, \label{eq:ss1}\\
q_i^t=1/{\left\| {(U_A^{t-1})^T x_i^A
-(U_B^{t-1})^T x_i^B )} \right\|_2}, \label{eq:ss2}\\
(U_A^{t-1},U_B^{t-1})= \arg \min _{U_I } J(U_I ,p_{ij}^{It} ,q^t_i),
\label{eq:ss3}
\end{eqnarray}
where $p_{ij}^I$ and $q_i$ are determined by the minimization
functions in half-quadratic minimization.
We can apply alternating minimization to (\ref{eq:ss3}). That is, we can fix
$U_B^{t-1}$ to find a solution of $U_A^{t}$ and then we make use of
$U_A^{t}$ to update $U_B^{t}$. Hence the solution of (\ref{eq:ss3})
can be obtained by minimizing the following two linear systems,
\begin{eqnarray}
U_A^{t} = X_A (I + \lambda _1 L^t_A  + \lambda _2 Q^t)X_A^T
\backslash (X_A \hat Y^t_A), \label{eq:su1}\\
U_B^{t} = X_B (I + \lambda _1 L^t_B  + \lambda _2 Q^t)X_B^T
\backslash (X_B \hat Y^t_B), \label{eq:su2}
\end{eqnarray}
where
\begin{eqnarray}
L^t_A =D^A-W^A,W^A_{ij}=w_{ij}p_{ij}^{At}, D^A_{ii}=\sum\nolimits_j {W_{ij}^A },\\
L^t_B =D^B-W^B,W^B_{ij}=w_{ij}p_{ij}^{Bt}, D^B_{ii}=\sum\nolimits_j {W_{ij}^B },\\
\hat Y^t_A= Y +\lambda _2 Q^t X_BU^{t-1}_B,
 \hat Y^t_B= Y +\lambda _2 Q^t X_AU^{t-1}_A,
\end{eqnarray}
where $D^A$ and $D^B$ are diagonal matrices. From the above three
equations, it can be seen that auxiliary variables $p^{It}$ and $q^t$
actually play a role of weighting to refine the structure in $z$ and
label matrix $Y$ during learning. This weighting strategy alleviates
the semantic gap problem and makes the proposed method more robust
to outliers. Algorithm~\ref{Alg:CMM} summarizes the above
optimization procedure.
\begin{algorithm}[]
\caption{Cross Modal Matching with Pairwise Constraint (CMMp)
\label{Alg:CMM}}
 \KwIn{$X_A \in R^{d1 \times n}$, $X_B \in R^{d2 \times n}$, $\lambda_1$ and $\lambda_2$.}
\begin{algorithmic}[1]
    \STATE Normalize samples in each $X_I$ to have unit $\ell_2$-norm.
    \STATE Compute semantic graph $Z$ via Algorithm \ref{Alg:CSC}.
    \REPEAT
    \STATE Compute auxiliary variables $p_{ij}^{It}$ and $q^t_i$ according to (\ref{eq:ss1}) and (\ref{eq:ss2}) respectively.
    \STATE Compute subspaces $U_A^{t}$ and $U_B^{t}$ according to (\ref{eq:su1}) and (\ref{eq:su2}) respectively.
    \STATE t=t+1.
    \UNTIL Converges
\end{algorithmic}
\end{algorithm}

According to the properties of convex functions, (\ref{eq:sobj}) is
joint convex so that there is a global minimum.
Proposition~\ref{th:scon} in Appendix~\ref{sec:app} ensures that
Algorithm \ref{Alg:CMM} converges to the global minimum.
The computational cost of Algorithm \ref{Alg:CMM} mainly involves
matrix multiplications and linear equation systems in (\ref{eq:su1})
and (\ref{eq:su2}), which can be efficiently solved by an iterative
algorithm LSQR \cite{Paige:1982}. Compared with eigen decomposition
methods~\cite{MKan:2012}\cite{Sharma:2012}, the computational costs
of linear equation systems tend to be very small \cite{Cai:2007}. In
addition, the empirical results in image processing, computer vision
and machine learning show that iteratively reweighted minimization
based methods often converge fast and only need a few iterations to
converge \cite{FNie:2010}\cite{RHe:2011}\cite{RHe:2012}.

\subsection{Relation to previous works \label{sec:rel}}
\subsubsection{Subspace clustering and cross-modal clustering} For
unsupervised learning, subspace clustering
\cite{Elhamifar:2013}\cite{GLiu:2010} has drawn much attention in
the computer vision community recently. A lot of efficient
subspace clustering algorithms
\cite{Elhamifar:2013}\cite{GLiu:2010}\cite{GLiu:2011}\cite{CLiu:2012}
have been developed. Recently, block-diagonal prior \cite{JFeng:2014}, smooth representation~\cite{HHu:2014}, and weight matrix based structure constraints \cite{KTang:2014} were introduced to further improve subspace clustering accuracy.
The proposed cross-modal subspace clustering
method in Algorithm \ref{Alg:CSC} is a natural extension of previous
single modality subspace clustering to multiple modalities.
Considering an ideal case of (\ref{eq:ucp}), i.e., $Z_A=Z_B=Z$, we have
\begin{equation}\label{eq:unucl}
\min _Z \sum\nolimits_I {w_I\left\| {X_I  - X_I Z} \right\|_F^2 }  +
\lambda \left\| {Z } \right\| \; s.t. \; diag(Z)=0,
\end{equation}
where $\left\|. \right\|$ is any matrix norm that has been used in
subspace clustering. We can further reformulate (\ref{eq:unucl}) as the following matrix trace minimization problem,
\begin{eqnarray} \label{eq:un_ss1}
\mathop {\min}\limits_Z tr((I-Z)^T (w_AX_A^TX_A + w_BX_B^TX_B)(I-Z)) \\
+ \lambda \left\| Z \right\| \quad s.t. \; diag(Z)=0, \nonumber
\end{eqnarray}
where $tr(.)$ is the matrix trace operator. Let
\begin{equation}
\hat{X} = \left[ {\begin{array}{*{20}c}
   {\sqrt {w_A }I_A } & 0  \\
   0 & {\sqrt {w_B }I_B }  \\
\end{array}} \right]\left[ {\begin{array}{*{20}c}
   {X_A }  \\
   {X_B }  \\
\end{array}} \right],
\end{equation}
where $I_A$ and $I_B$ are identity matrices. Then (\ref{eq:unucl}) and (\ref{eq:un_ss1}) take the following form,
\begin{equation} \label{eq:un_ss}
\min _Z\left\| {\hat{X}  - \hat{X} Z} \right\|_F^2   +
\lambda \left\| {Z } \right\| \; s.t. \; diag(Z)=0.
\end{equation}
It is interesting to observe that (\ref{eq:un_ss}) is a standard formulation in subspace
clustering and can be solved by the standard solvers~\cite{Elhamifar:2013}\cite{GLiu:2010}. The problem in (\ref{eq:un_ss}) can be viewed as the naive method to concatenate the feature vectors in each modality.

In cross-modal learning, it has been demonstrated that the simple concatenation of multi-modal feature vectors will not improve accuracy so much. For subspace clustering, although the naive method in (\ref{eq:un_ss}) has another formulation in (\ref{eq:unucl}) assuming $Z_A=Z_B=Z$, it still does not work well due to the semantic gap between different modalities. Since the representation abilities of multi-modal features are imbalanced, it is difficult to use a unique subspace $Z$ to represent different modalities. To alleviate this problem, we only assume in (\ref{eq:ucp}) that the subspace representation $Z_I$ of each modality should be close to each other. Experimental results in Section~\ref{sec:expu} demonstrate that the proposed model in (\ref{eq:ucp}) can alleviate this problem and further improve clustering accuracy.


Previous multi-view spectral clustering methods
\cite{Kumar:2011}\cite{Kumar:2011_2} and deep network based methods
\cite{YKang:2012}\cite{Srivastava:2012} try to learn common
representations before clustering. However, our proposed
method is derived from pairwise constraint and aims to learn a
shared structure from different modalities. It
can be viewed as a variant of graph based multi-modality learning
\cite{HTong:2005}. Different from \cite{HTong:2005}, our method resorts to subspace
clustering to learn a common graph rather than fusion of the graphs
from different modalities \cite{HTong:2005}. Compared with
multi-task clustering \cite{QGu:2009}\cite{JHe:2012}, there is only
one task between different modalities. Recently, wang et al.
\cite{HWang:2013} applied structured sparsity to learn cluster
indication matrix and then used K-means to perform multi-view
clustering. Different from \cite{HWang:2013}, our cross-modal
subspace clustering method in Algorithm~\ref{Alg:CSC} applies the
recent linear representation based subspace clustering technique.

\subsubsection{Cross-modal Retrieval} For supervised
learning, our proposed cross-modal matching method has a close
relationship to graph embedding based methods. Because of the linear
regression formulation of graph embedding
\cite{Ye:2007}\cite{Cai:2007}, our method can be viewed as a
multi-modal extension and combination of LDA and CCA. It keeps
intra-class variation like LDA meanwhile handles pairwise constraint
like CCA. Different from common discriminant feature extraction
(CDFE) \cite{DLin:2006}, the supervised version of
CCA~\cite{Kim:2007}, locality preserving CCA \cite{TSun:2007} and
multi-view LDA \cite{MKan:2012}, the proposed method can handle
intra-class variation, pairwise constraint and structure preserving
at the same time. In particular, we preserve the same semantic
structure for different modalities rather than two structures for
two modalities respectively in
\cite{DLin:2006}\cite{TSun:2007}\cite{MKan:2012}. In addition, the
proposed method is robust to the outliers in pairwise constraints
due to its $\ell_{21}$-norms.
Inspired by the bilinear model in \cite{Tenenbaum:2000}, Sharma et
al. \cite{Sharma:2012} extend graph embedding framework
\cite{Yan:2007} to multi-modal learning that aims to solve the following
eigen decomposition problem,

\begin{table*}[htbp]
\begin{center}
\begin{tabular}{|c||c|c|c||c|c|c|}
\hline
\multirow{2}{*}{Methods} & \multicolumn{3}{|c||}{Accuracy (\%)} & \multicolumn{3}{|c|}{Normalized Mutual Information (\%)}  \\
\cline{2-7}
& Wiki & VOC & Digits  & Wiki & VOC & Digits \\
\hline
Spectral\_S  & 52.96 $\pm$ 2.65 & 80.89 $\pm$ 3.13 & 68.50 $\pm$ 5.06 & 55.96 $\pm$ 1.92 & 61.75 $\pm$ 3.40 & 64.71 $\pm$ 2.27  \\
\hline
Spectral\_M  & 55.44 $\pm$ 2.15 &84.88 $\pm$ 0.86 & 74.15 $\pm$ 4.73  & 54.95 $\pm$ 0.88 & 66.53 $\pm$ 1.32 & 71.69 $\pm$ 1.80  \\
\hline
Bipartite    & 55.57 $\pm$ 1.87 &76.37 $\pm$ 4.13 & 76.93 $\pm$ 4.12  & 55.66 $\pm$ 1.31 & 56.74 $\pm$ 4.13 & 73.79 $\pm$ 1.29  \\
\hline
Co\_Pairwise & 55.63 $\pm$ 1.49 &82.51 $\pm$ 0.00& 81.31 $\pm$ 5.54   & 54.28 $\pm$ 1.83 & 63.35 $\pm$ 0.00 & 76.98 $\pm$ 2.30 \\
\hline
Co\_Centroid & 56.47 $\pm$ 1.86 &79.34 $\pm$ 0.69& 81.39 $\pm$ 3.41   & 56.75 $\pm$ 0.61 & 59.03 $\pm$ 0.00 & 75.34 $\pm$ 2.73   \\
\hline
Co\_Training & 56.34 $\pm$ 1.95 &84.88 $\pm$ 0.00& 81.47 $\pm$ 4.59   & 56.46 $\pm$ 0.66 & 63.17 $\pm$ 0.00 & 75.07 $\pm$ 2.04  \\
\hline
Multi\_NMF   & 56.07 $\pm$ 2.29 &84.65 $\pm$ 3.89 & --- ---           & 56.92 $\pm$ 1.05 &67.21 $\pm$ 3.81 & --- ---  \\
\hline
Multi\_CF    & 59.87 $\pm$ 2.96 &92.58 $\pm$ 6.27 & 81.01 $\pm$ 9.24  & 57.64 $\pm$ 0.63 & 75.06 $\pm$ 3.81 & 80.05 $\pm$ 4.67  \\
\hline
LSR\_S       & 53.04 $\pm$ 2.94 &92.13 $\pm$ 0.11 & 78.23 $\pm$ 1.59  & 56.32 $\pm$ 1.59 & 68.17 $\pm$ 0.28 & 74.89 $\pm$ 1.03  \\
\hline
LSR\_M       & 56.28 $\pm$ 1.96 &95.81 $\pm$ 0.00 & 85.20 $\pm$ 5.92  & 55.06 $\pm$ 1.18 &75.26 $\pm$ 0.00 & 71.72 $\pm$ 1.83  \\
\hline
CSC          & \textbf{61.48 $\pm$ 1.25}  &\textbf{96.54 $\pm$ 0.00} & \textbf{88.52 $\pm$ 3.09}  & \textbf{58.76 $\pm$ 2.24} & \textbf{85.15 $\pm$ 0.00} & \textbf{83.80 $\pm$ 2.36} \\
\hline
\end{tabular}
\end{center}
\caption{Clustering results of different methods over 20 runs.}
\label{result}
\end{table*}

\begin{eqnarray} \label{eq:GM}
&\mathop {\max }\limits_{u_A ,u_B } u_A^T M_A u_A  + u_B^T M_B u_B  \\
&s.t. \; u_A^T K_A u_A  + u_B^T K_B u_B  = 1, \nonumber
\end{eqnarray}
where $M_A$ and $M_B$ are some symmetric square matrices and $K_A$ and $K_B$ are square
symmetric definite matrices. The values of $M_A$, $M_B$, $K_A$ and $K_B$ can be specified
according to a graph embedding method. However, the bilinear model
in \cite{Tenenbaum:2000} does not deal with pairwise constraint
during learning, and the generalized multiview analysis in
\cite{Sharma:2012} approximates pairwise constraint by making the
multi-view samples within the same class. Both of them do not
efficiently make use of pairwise constraint during learning, which
is an important issue in web documents. Our proposed method is also
different from the methods for image annotations in
\cite{Weston:2010}\cite{Lucchi:2012} due to the fact that the
semantic gap between texts and images in web documents is larger
than that between images and their annotations.

\section{Experiments \label{sec:exp}}
In this section, we apply our proposed unsupervised and supervised multi-modal learning methods
to clustering and retrieval tasks respectively. For a
fair evaluation, all results are averaged over 20 independent runs,
with the mean error and standard deviation reported.

\subsection{Cross-modal Clustering\label{sec:expu}}
\subsubsection{Algorithms}
To evaluate the clustering performance of the proposed cross-modal subspace clustering (\textbf{CSC}) method, we compare our CSC method with the following algorithms.

\textbf{Spectral\_S:} The spectral clustering method~\cite{Ncuts} is used to cluster each modalities and the best result is reported.

\textbf{Spectral\_M:} The spectral clustering method of \cite{Ncuts} is used to perform clustering on the concatenated features of all modalities.

\textbf{Bipartite:} A bipartite graph \cite{Bipartite} is constructed from two modalities, and then a standard spectral clustering method is used to cluster data.

\textbf{Co\_Pairwise, Co\_Centroid\footnote{http://www.umiacs.umd.edu/~abhishek/papers.html \label{fn1:repeat1}}}\textbf{:}
Two co-regularization methods on the eigenvectors of the Laplacian matrices from all modalities~\cite{Kumar:2011_2}.

\textbf{Co\_Training\footref{fn1:repeat1}}\textbf{:} Alternately modifying one modality's graph structure using the other modality's information \cite{Kumar:2011}.

\textbf{Multi\_NMF\footnote{http://jialu.cs.illinois.edu/publication}}: A multi-modal non-negative matrix factorization method to group the database~\cite{Multi-NMF}.
Since Multi\_NMF requires that the feature matrix should be
non-negative, we only report its results on the dataset with non-negative features.

\textbf{Multi\_CF:} A structure sparsity based multi-modal clustering and feature learning framework~\cite{HWang:2013}.


\textbf{LSR\_S} The subspace clustering via least squares regression \cite{CLiu:2012} is used to cluster each modality's data and the best result is reported.

\textbf{LSR\_M:} The subspace clustering via least squares regression \cite{CLiu:2012} is used to perform clustering on the concatenated features of all modalities.

Two commonly used measures, clustering accuracy and normalized mutual
information (NMI) \cite{NMI}, are used to measure clustering results.
For the methods that apply Gaussian kernel to construct an affinity matrix,
the Gaussian kernel size parameter is determined by the mean value of the Euclidean
distance between all data points. For Co\_Pairwise, Co\_Centroid,
Co\_Training, Multi\_NMF and Multi\_CF methods, we follow the suggestions of the authors to
achieve their best clustering results. For our proposed CSC method, we simply
make $\omega_I$ the same for all modalities because there is no prior
knowledge. $\lambda_1$ and $\lambda_3$ are empirically set to
reach the best clustering performance.

\subsubsection{Databases}
As operated in \cite{Kumar:2011_2}\cite{Multi-NMF}, three public datasets are used to evaluate the clustering performance. The settings of these datasets are as follows,

Wiki Text-image dataset \cite{Rasiwasia:2010}\cite{YChen:2012}\cite{Sharma:2012} consists of 2173/693 (training/ testing) image-text pairs
from 10 semantic classes. It has a 10 dimensional latent
Dirichlet allocation model based text features and 128 dimensional SIFT
histogram image features. Since the number of training samples of each class is different,
we randomly select 60 samples per class from the Wiki training dataset to evaluate different clustering methods.

Pascal VOC 2007 dataset\footnote{http://pascallin.ecs.soton.ac.uk/challenges/VOC/voc2007/} consists of
20 categories, including 5,011 training and 4,952 testing image-tag pairs. GIST features are used for
the images and word frequency features are used for tags. Some of the pairs are multi-labeled, so we
only select those with one label. Besides, those tag features with only zeros are also removed.
Finally, the first three categories are selected as a subset to evaluate different clustering methods.

UCI Handwritten Digit dataset\footnote{http://archive.ics.uci.edu/ml/datasets/Multiple+Features} is composed of
multi-modal features of handwritten numerals (0--9), which are extracted from a collection of Dutch utility maps.
It consists 10 categories, each of which has 200 samples. We select 76 Fourier coefficients of the character shapes
and 64 Karhunen-Love coefficients as the two modalities of the original dataset.

For each dataset, we normalized each sample to have unit $\ell _2$-norm for all compared algorithms. On the Wiki Text-image dataset, we perform the random selection for 20 times and report average results. On the left two dataset, we repeat each clustering algorithm for 20 times on one selected dataset.


\subsubsection{Numerical results}
Table \ref{result} tabulates the clustering results of different clustering algorithms on the three
public datasets. We observe that cross-modal clustering methods perform better than single-modal
methods, which indicates that each modality's data are helpful for clustering. Our proposed CSC
method performs better than its competitors in terms of both clustering accuracy
and normalized mutual information. Subspace clustering methods (including LSR\_S, LSR\_M, CSC) seem to be more suitable for clustering
tasks on the VOC dataset. This may be because they model the structure of data more accurately.


%

Experimental results also show that although the discriminative ability of different modalities
is different, different modalities are complementary for each other. Comparing Spectral\_S, Spectral\_M,
LSR\_S and LSR\_M, we observe that by just concatenating features of all modalities, traditional single
modality method can obtain at least 3\%-5\% improvement in terms of clustering accuracy. The clustering
accuracy improvements of LSR\_M over LSR\_S are 3.68\% and 6.97\% on the two datasets respectively. These
improvements indicate that clustering performance can be further improved if the two modalities are well used.
Although LSR\_M, LSR\_S and CSC all apply subspace clustering technique to deal with multi-modal problems,
our proposed CSC method provides an efficient way to deal with pairwise constraints so that
it can better exploit the complementariness of multiple modalities and achieves the best results.

\begin{figure}[t]
\begin{center}
    \subfigure[]{\includegraphics[width=0.78\linewidth]{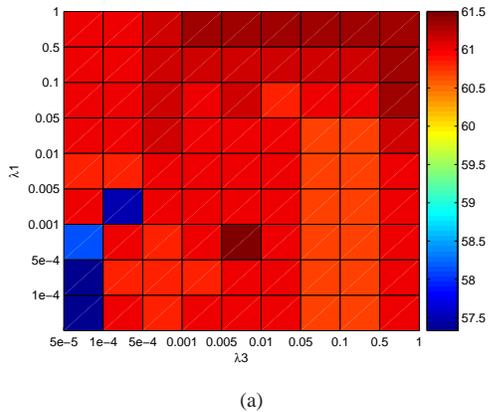}}
\end{center}
   \caption{The clustering accuracy of CSC under different parameter
   settings on the Wiki Text-image dataset. \label{fig:CSCParam}}
\end{figure}

\subsubsection{The parameter setting of CSC}
For the proposed CSC method, $\lambda_1$ and $\lambda_3$ control the prior structure
on subspace representations and the pairwise constraints on different modalities respectively.
Fig.~\ref{fig:CSCParam} shows the clustering accuracy as a function of $\lambda_1$ and $\lambda_3$. The experimental setting is the same
as that in the Wiki Text-image dataset. We observe that both of these two parameters are important. More important, there is a large range for $\lambda_1$
and $\lambda_3$ to make CSC outperform its competitors. The pair-wise constraint corresponding to the regularization $\left\|Z_A-Z_B\right\|$ plays an important
role in CSC. It makes the subspace representations of different modalities close to each
other, which potentially leads to an improvement in clustering accuracy. In addition,
$\omega_I$ balances the importance of each modality. In our experiments, we simply fix them
to be the reciprocal of the number of modalities.

\begin{table*}
\begin{center}
\begin{tabular}{|l|c|c|c|c|c|c|c|c|c|c|c|}
\hline
Dataset & PCA &LDA & BLM & CCA & LPCCA & PLS  & CDFE & SliM{$^{2}$} & GMLDA & CMMp \\
\hline\hline
Tr(70)      & 0.131 & 0.131 & 0.134 & 0.165 & 0.171 & 0.176  & 0.174 & 0.187 & 0.199 & \textbf{0.228}  \\
Tr(100)     & 0.132 & 0.130 & 0.135 & 0.174 & 0.178 & 0.180 & 0.182  & 0.193 & 0.201 & \textbf{0.233}  \\
Tr(130)     & 0.132 & 0.130 & 0.135 & 0.179 & 0.181 & 0.173 & 0.190  & 0.194 & 0.203 & \textbf{0.236}  \\
\hline
\end{tabular}
\end{center}
\caption{Average MAP scores for text query of different methods over
20 runs. Notations 'Tr(70)', 'Tr(100)' and 'Tr(130)' indicate that
70, 100, and 130 samples per class are randomly selected from the
Wiki training dataset for training respectively. \label{tab:smap}}
\end{table*}
\begin{table*}
\begin{center}
\begin{tabular}{|l|c|c|c|c|c|c|c|c|c|c|}
\hline
Dataset & PCA & LDA & BLM & CCA & LPCCA & PLS & CDFE & SliM{$^2$} & GMLDA & CMMp \\
\hline\hline
Tr(70)      &  9.5$\pm$5.2 & 10.6$\pm$2.7 & 12.2$\pm$4.2 & 33.2$\pm$4.9 & 35.8$\pm$4.3 & 25.8$\pm$4.0  & 34.7$\pm$6.4 & 36.9$\pm$3.6 & 21.9$\pm$4.7 & \textbf{41.2}$\pm$5.4 \\
Tr(100)     & 16.2$\pm$5.4 &  9.9$\pm$5.2 & 13.5$\pm$7.6 & 39.4$\pm$5.4 & 38.7$\pm$4.6 & 32.5$\pm$5.6 & 37.1$\pm$3.7  & 40.8$\pm$4.2 & 19.6$\pm$4.0 & \textbf{44.0}$\pm$5.2  \\
Tr(130)     & 12.6$\pm$3.1 & 11.1$\pm$4.7 & 15.9$\pm$8.0 & 41.5$\pm$5.5 & 39.9$\pm$3.5 & 36.6$\pm$3.5 & 40.7$\pm$6.5  & 43.1$\pm$3.3 & 22.8$\pm$2.5 & \textbf{47.6}$\pm$4.8  \\
\hline
\end{tabular}
\end{center}
\caption{Average recognition rates and standard deviations for text
query by using the KNN (K=10) classifier. \label{tab:srec}}
\end{table*}

\subsection{Cross-modal retrieval \label{sec:exps}}
\subsubsection{Algorithms}
In this subsection, we make use of principal component analysis
(PCA), linear discriminant analysis (LDA), canonical correlational
analysis (CCA) \cite{Rasiwasia:2010}, and partial least squares
(PLS)
\cite{Sharma:2011}\cite{YChen:2012}\footnote{http://www.cs.umd.edu/~djacobs/pubs\_files/PLS\_Bases.m}
as the baselines for cross-modal retrieval. We also compare five
cross-modal learning methods, including bilinear model (BLM) for
multi-view learning \cite{Tenenbaum:2000}, common discriminant
feature extraction (CDFE) \cite{DLin:2006}, locality preserving CCA
(LPCCA) \cite{TSun:2007}, SliM$^2$ \cite{YZhuang:2013}, and generalized multi-view linear
discriminant analysis (GMLDA)
\cite{Sharma:2012}\footnote{http://www.cs.umd.edu/~bhokaal/Research.htm}.
As reported in \cite{Sharma:2012}, GMLDA often achieves the highest
MAP. Hence we only discuss GMLDA in this section.

Since the number of training samples of each category is different,
we randomly select 70, 100 and 130 samples per class from the Wiki
training dataset as three training sets respectively. We
make use of the Wiki testing dataset as our testing set. Hence the
used training and testing sets are different. The parameters of all
compared methods are empirically tuned to achieve the best results,
and all results are averaged over 20 independent runs.  Mean average
precision (MAP) and recognition rate are used as the evaluation
criterion and $\ell_2$ distance is used as the distance function.
For MAP\footnote{http://pascallin.ecs.soton.ac.uk/challenges/VOC/},
precision at 11 different recall levels \{0, 0.1, 0.2, 0.3, 0.4,
0.5, 0.6, 0.7, 0.8, 0.9, 1.0\} is used as in \cite{Sharma:2012}; and
for recognition rate, we use the $K$ nearest neighbors (KNN)
classifier. Since there is ten classes, we set $K$ to 10.

\begin{figure*}[t]
\begin{center}
   \includegraphics[width=0.98\linewidth]{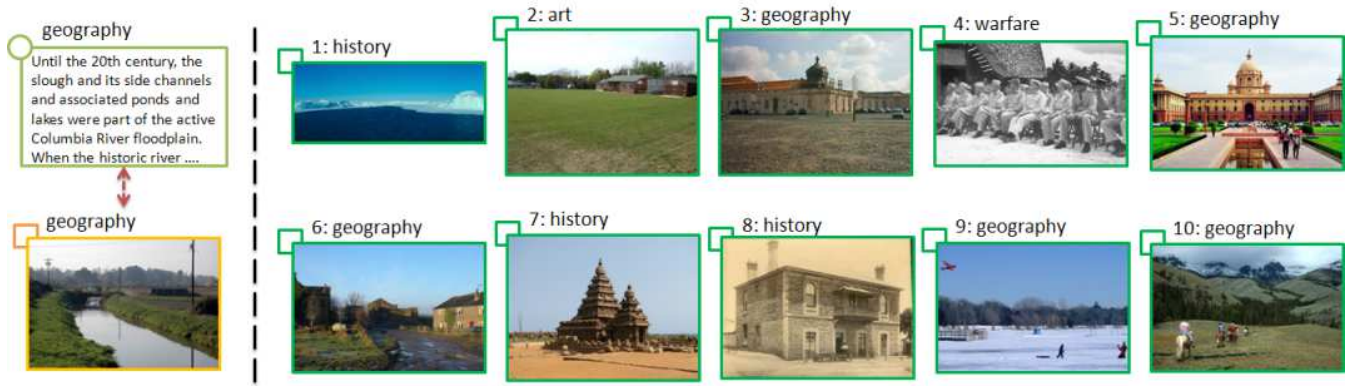}
\end{center}
   \caption{Example of a text query and its corresponding top retrieved images. The left column
   contains the query text object and its paired image; and the right column lists the top ten retrieved images. \label{fig:illu}}
\end{figure*}

\subsubsection{Numerical results on the Wiki dataset}
The commonly used Wiki text-image dataset
\cite{Rasiwasia:2010}\cite{YChen:2012}\cite{Sharma:2012} is used
to highlight the benefits of the pairwise constraint. Wiki Text-image
dataset consists of 2173/693 (training/ testing) image-text pairs
from 10 different semantic classes. It has a 10 dimensional latent
Dirichlet allocation model based text features and 128 dimensional SIFT
histogram image features.

Table \ref{tab:smap} tabulates average MAP scores
for the text query over 20 runs. We see that four baseline methods
can be ordered in ascending MAP scores as LDA, PCA, CCA, PLS. Since
LDA and PCA do not consider the correlation between different
modalities, they fail in cross-modal tasks. It is clear that LPCCA,
CDFE, SliM$^2$, GMLDA and CMMp perform better than LDA and CCA because they
can better use structure information from different modalities.
Compared with LPCCA, CDFE, SliM$^2$, and GMLDA, our CMMp can handle intra-class
variation, pairwise constraint and structure preserving in a
framework such that it achieves the highest average MAP scores in
all cases.

To further demonstrate the effectiveness of CMMp for cross-modal
retrieval, we list recognition rates for text query in Table
\ref{tab:srec}. A higher recognition rate indicates that the
retrieved top ten images contain more corrected images belonging to
the same category of the query text. It is interesting to see that
although CDFE and GMLDA have higher MAP scores than CCA, they have
lower recognition rates than CCA. This indicates that CDFE and GMLDA
obtain a better overall rank than CCA. However, if users only focus
on top ten retrieved results, CDFE and GMLDA may perform worse than
CCA. Our proposed CMMp achieves the best results in terms of both
recognition rates and MAP scores, which also demonstrates the
proposed method is potentially powerful for cross-modal learning.

Fig. \ref{fig:illu} depicts an example of a text query and its
corresponding top retrieved images. We can see that although the
category of the text query belongs to 'geography', its paired image
may be more similar to those images in categories 'history' and
'art' in the image space. However, five images from 'geography' are
still retrieved by our CMMp method. This may be because our CMMp
method can handle both intra-class variation and pairwise
constraint.

\begin{table}
\begin{center}
\begin{tabular}{|l|c|c|c|c|c|}
\hline
\small{Dataset} & $\ell_1$ & cosine & ChiSq &\small{SCM\cite{Rasiwasia:2010}}&\small{GMA\cite{Sharma:2012}}\\
\hline\hline
\small{Tr(70)}  & 0.227 & 0.265 & 0.152 & 0.226 &0.232\\
\small{Tr(100)} & 0.233 & 0.271 & 0.153 & 0.226 &0.232\\
\small{Tr(130)} & 0.237 & 0.278 & 0.154 & 0.226 &0.232\\
\hline
\end{tabular}
\end{center}
\caption{Average MAP scores of CMMp for text query under various
distance functions over 20 runs. \label{tab:df}}
\end{table}

\begin{table*}
\begin{center}
\begin{tabular}{|l|c|c|c|c|c|c|c|c|c|c|}
\hline
Dataset & PCA &PCA+LDA & BLM & CCA & LPCCA & PLS  & CDFE &SliM$^2$& GMLDA & CMMp \\
\hline\hline
Text query     & 0.092 & 0.122 & 0.115 & 0.132 & 0.074 & 0.144 & 0.071 & 0.154 & 0.170 & \textbf{0.171}  \\
Image query    & 0.058 & 0.111 & 0.082 & 0.103 & 0.087 & 0.110 & 0.084 & 0.167 & 0.100 & \textbf{0.170}  \\
Average        & 0.075 & 0.117 & 0.099 & 0.118 & 0.081 & 0.127 & 0.078 & 0.161 & 0.135 & \textbf{0.171}  \\
\hline
\end{tabular}
\end{center}
\caption{MAP scores of different methods on the VOC dataset.
\label{tab:sVOC}}
\begin{center}
\begin{tabular}{|l|c|c|c|c|c|c|c|c|}
\hline
Dataset & LDA & BLM & CCA & LPCCA & PLS  & CDFE & GMLDA & CMMp \\
\hline\hline
Text query      & 0.122 & 0.172 &0.200 &0.075 & 0.181 & 0.201 & 0.237 & \textbf{0.245}  \\
Image query     & 0.111 & 0.118 &0.165 &0.068 & 0.156 & 0.163 & 0.179 & \textbf{0.213}  \\
Average         & 0.117 & 0.145 &0.183 &0.072 &0.169 & 0.182  & 0.208 & \textbf{0.229} \\
\hline
\end{tabular}
\end{center}
\caption{MAP scores of different methods on the VOC dataset. PCA is
used as a preprocessing step for all methods. \label{tab:sVOCp}}
\end{table*}
\begin{figure*}[t]
\begin{center}
   \subfigure[]{\includegraphics[height=46mm,width=55mm]{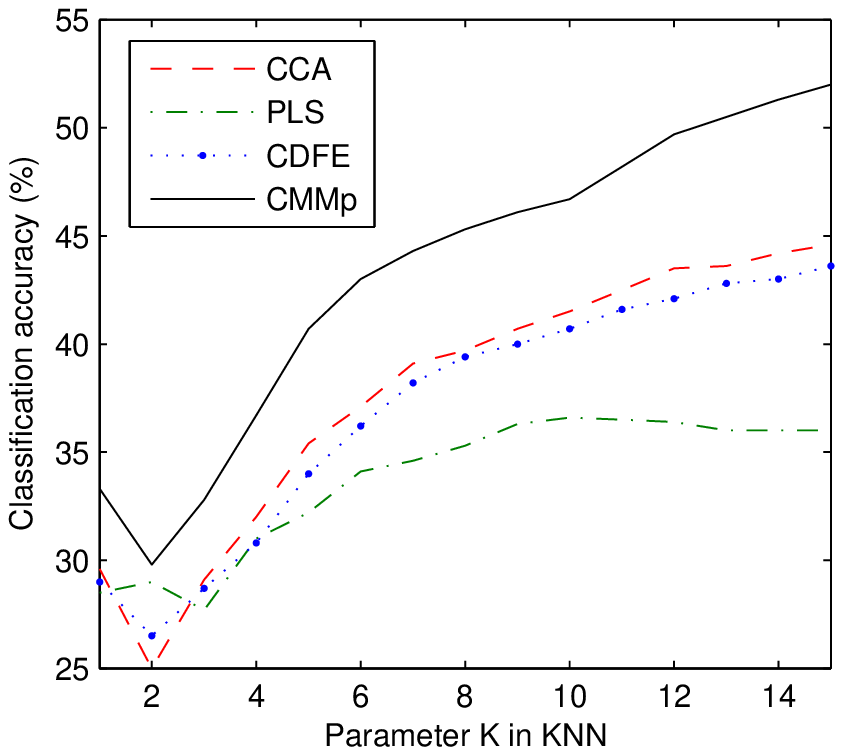}}
   \subfigure[]{\includegraphics[height=46mm,width=55mm]{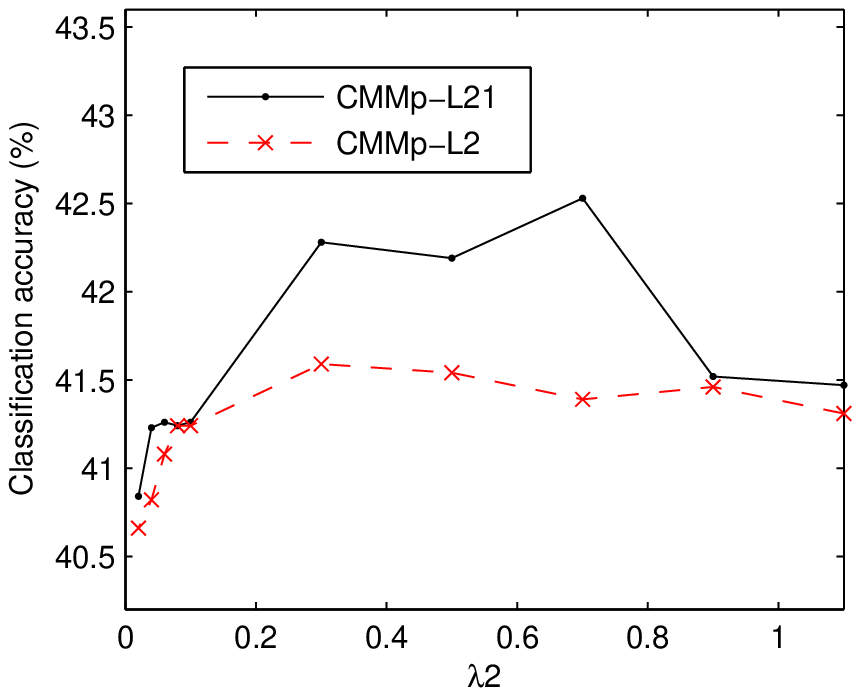}}
   \subfigure[]{\includegraphics[height=46mm]{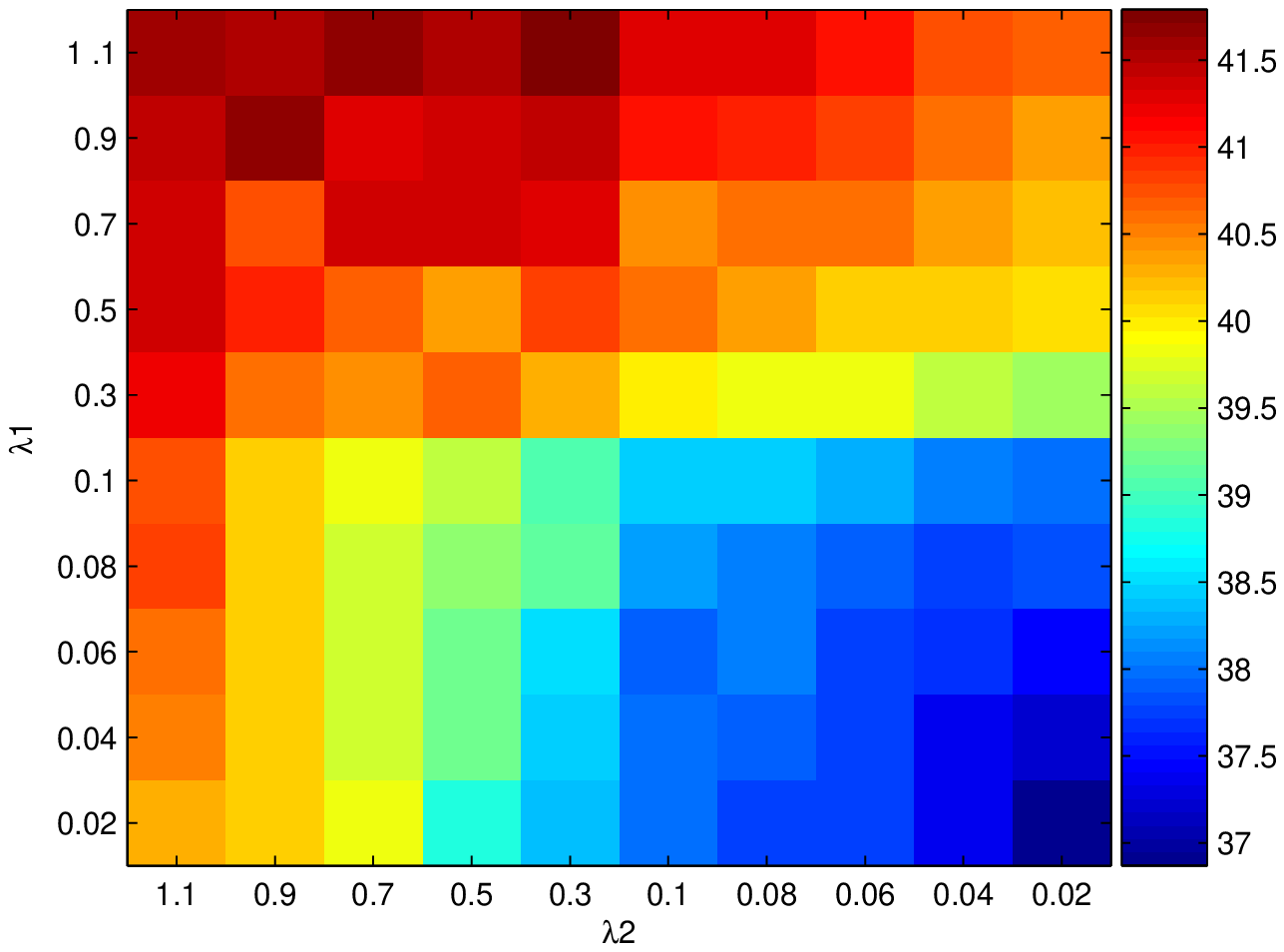}}
\end{center}
   \caption{The performance of CMMp under different parameter settings.
   (a) Classification accuracy as a function of $K$ in KNN classifier.
   (b) Classification accuracy as a function of $\lambda_2$ in CMMp.
   CMMp-L21 and CMMp-L2 indicate that $\ell_{21}$ norm and $\ell_2$
   norm are used in the last item in (\ref{eq:sobj}) respectively.
   (c) Classification accuracy as a function of $\lambda_1$ and $\lambda_2$ in CMMp.
   \label{fig:sparam}}
\end{figure*}

Table \ref{tab:df} gives MAP scores under various distance
functions\footnote{http://www.cs.columbia.edu/~mmerler/project/code/pdist2.m}.
Since semantic correlation matching (SCM) with a linear kernel
\cite{Rasiwasia:2010} and generalized multiview analysis (GMA)
\cite{Sharma:2012} have shown the state-of-the-art performance for the
Wiki text-image dataset, we also report the best results in
\cite{Rasiwasia:2010}\cite{Sharma:2012}. As in
\cite{Rasiwasia:2010}, $\ell_1$ distance and $\ell_2$ distance lead
to similar MAP scores. In particular, the cosine distance can
significantly improve MAP scores. When we increase the number of
training samples in the training, the MAP scores of our proposed
method are better than the best results reported in
\cite{Rasiwasia:2010}\cite{Sharma:2012} if $\ell_2$ and $\ell_1$
distances are used. Table \ref{tab:df} also demonstrates that the
MAP scores of our proposed method can be further improved if a
suitable distance function is adopted.

\subsubsection{Experimental results on the VOC dataset}
To further
evaluate different cross-modal matching methods, we perform
experiments on a subset of Pascal VOC
2007 \cite{HWang:2011}, which consists of collected
5011/4952 (training/testing) image-tag pairs belonging to 20
different categories. We make use of 512-dimensional Gist features
and 399-dimensional word frequency features for image and tag
respectively. Since there are zero vectors and multi-labeled images,
we select the images with only one object from the training and
testing set as in \cite{Sharma:2012}. As a result, we obtain 2799
training and 2820 testing data that correspond to 20 classes.

Tables \ref{tab:sVOC} and \ref{tab:sVOCp} show MAP scores of
different methods on the VOC dataset without and with PCA as a
preprocessing step respectively. We see that when PCA is used as a
preprocessing step to remove useless information, MAP scores of
almost all cross-modal methods are significantly improved. GMLDA
and CMMp perform better than other methods. This may be because they
can handle both discriminative and cross-modal information. Since
CMMp applies $\ell_{21}$ norms to deal with inaccurate pairs from
two modalities, it achieves higher MAP scores than GMLDA.

The imbalance of different modalities and diverse description of image modality make cross modal retrieval more challenging. For example, the first, second and sixth images in Fig. \ref{fig:illu} belong to 'history', 'art' and 'geography' categories respectively. However, without any prior knowledge, one may classify all the three images into the 'geography' category. Compared with image modality, text modality has a narrative (or specific) description \cite{YJia:2011}, which makes text modality be more discriminative. On the VOC dataset (Table \ref{tab:sVOCp}), the highest MAP scores for text query and image query are 0.245 and 0.213 respectively. An important issue for cross modal retrieval may be to balance the narrow description of text modality and the diverse description of image modality. A potential solution to this issue may be the combination of feature selection to select most relevant image features or regions to narrow the diverse description of image modality.

\subsubsection{The parameter setting of CMMp} The regularization
parameters in graph embedding based subspace methods often
significantly affect the classification accuracy. In this section,
we discuss the parameter setting of our proposed cross-modal
matching methods.

Classification accuracy as a function of $K$ in KNN classifier is
given in Fig. \ref{fig:sparam} (a). We observe that classification
rates of all methods increase quickly as $K$ increases. This may be
because more retrieved images corresponding to the input category
are selected when $K$ increases. Our CMMp can achieve higher
classification rates than the other three methods, which indicates
that our method can select more correct images than the other three
methods in retrieved top $K$ images. Fig. \ref{fig:sparam} (a) also
gives an explanation that CMMp can obtain better MAP results.

Classification accuracy as a function of $\lambda_2$ in CMMp is
plotted in Fig. \ref{fig:sparam} (b). Here we further discuss two
regularizers. CMMp-L21 and CMMp-L2 indicate that $\ell_{21}$ norm
and $\ell_2$ norm are used in the last item in (\ref{eq:sobj})
respectively. We observe that CMMp-L21 consistently performs better
than CMMp-L2, which indicates the $\ell_{21}$-norm in the last item
is necessary. This may be due to outliers in the pairwise constraint
of web documents. Image features are often less discriminative than
text features such that some text-image pairs are inaccurate.

Classification accuracy as a function of $\lambda_1$ and $\lambda_2$
in CMMp is shown in Fig. \ref{fig:sparam} (c). We see that the
setting of $\lambda_1$ and $\lambda_2$ will affect the
classification accuracy. The highest classification rate is achieved
when both $\lambda_1$ and $\lambda_2$ are set to be larger than 1.
When one of $\lambda_1$ and $\lambda_2$ is set to a smaller value,
classification rates decrease. The variation of classification
accuracy indicates that the last two items in (\ref{eq:sobj}) play
an important role and handle structure preserving and pairwise
constraint respectively.

\section{Conclusion and future work \label{sec:con}}
This paper has systematically studied the pairwise constraint
problems in cross-modal learning, and has proposed a general
regularization framework for developing cross-modal learning
algorithms. For unsupervised learning, a cross-modal subspace
clustering method has been proposed to learn a common structure for
different modalities; and for supervised learning, a cross-modal
matching method has been proposed for multimedia retrieval.
Extensive experiments on the Wiki and VOC datasets demonstrate
that the joint text and image modeling with pairwise constraint can
improve clustering or matching accuracy.
In the future, one potential direction is to apply the proposed
framework in (\ref{eq:prob}) to discriminative dictionary learning
with paired samples \cite{HGuo:2012}\cite{KJia:2013}. Another direction may be to narrow the diverse description of image modality by combining coupled feature selection in (\ref{eq:prob}) to select most relevant image features or regions, or using deep learning to learn more discriminative and related feature representations.


%

\appendices
\section{The Convergence of Algorithm~\ref{Alg:CMM}\label{sec:app}}
\newtheorem{proposition}{Proposition}
\begin{proposition}\label{th:scon}
Algorithm \ref{Alg:CMM} monotonically decreases the objective
function in (\ref{eq:sobj}) in each iteration, and converges to the global
optimum.
\begin{proof}
According to the properties of the minimization function in half-quadratic optimization~\cite{FNie:2010}\cite{RHe:2012}, we have,
$$J(U_I^{t-1} ,p^{I(t)} ,q^{t}) \le J(U_I^{t-1} ,p^{I(t-1)} ,q^{t-1}).$$
And according to (\ref{eq:ss3}), we obtain,
$$J(U_I^{t} ,p^{I(t)} ,q^{t}) \le J(U_I^{t-1} ,p^{I(t)} ,q^{t}) \le J(U_I^{t-1} ,p^{I(t-1)} ,q^{t-1}).$$
Therefore, Algorithm \ref{Alg:CMM} monotonically decreases the
objective function in (\ref{eq:sobj}).

In addition, for convex functions $f(x)$ and $g(x)$, $h(x)=f(x)+g(x)$ and $h(x)=f(Dx+b)$
are also convex functions\footnote{http://en.wikipedia.org/wiki/Convex\_function},
where $x \in R^{2n}$, $D \in R^{2n \times 2n}$, and $b \in R^{2n}$. Since we can reformulate
$\left\| {X_A^T U_A  - X_B^T U_B }\right\|_{21}$ as an affine map ($h(x)=f(Dx)$),
(\ref{eq:sobj}) is joint convex with respect to $U_A$ and $U_B$. Taking the derivative
of (\ref{eq:sobj}) w.r.t $U_A$ and $U_B$, and setting the derivative to zero, we arrive at:
\begin{eqnarray}
U_A = X_A (I + \lambda _1 L_A  + \lambda _2 Q)X_A^T
\backslash (X_A \hat Y_A), \label{eq:con1}\\
U_B = X_B (I + \lambda _1 L_B  + \lambda _2 Q)X_B^T
\backslash (X_B \hat Y_B), \label{eq:con2}
\end{eqnarray}
where
\begin{eqnarray}
L_A =D^A-W^A,W^A_{ij}=w_{ij}p_{ij}^{A}, D^A_{ii}=\sum\nolimits_j {W_{ij}^A }, \nonumber\\
L_B =D^B-W^B,W^B_{ij}=w_{ij}p_{ij}^{B}, D^B_{ii}=\sum\nolimits_j {W_{ij}^B }, \nonumber\\
\hat Y_A= Y +\lambda _2 Q X_B U_B,
\hat Y_B= Y +\lambda _2 Q X_A U_A, \nonumber\\
p^{I}_{ij}=1/{\left\| {U_I^T (x_i^I
-x_j^I )} \right\|_2}, \nonumber \\
Q_{ii}=q_i=1/{\left\| {U_A^T x_i^A
-U_B^T x_i^B )} \right\|_2}.  \nonumber
\end{eqnarray}
Since the problem in (\ref{eq:sobj}) is a joint convex problem, $U_A$ and $U_B$ are a global optimum solution to the problem
if and only if (\ref{eq:con1}) and (\ref{eq:con2}) are satisfied.

Since Algorithm \ref{Alg:CMM} will monotonically decrease the objective function in (\ref{eq:sobj}) in each iteration $t$, $U_I^t$, $p^{It}$ and $q^t$ will satisfy (\ref{eq:con1}) and (\ref{eq:con2}) in the convergence. As the problem in (\ref{eq:sobj}) is a joint convex problem, satisfying (\ref{eq:con1}) and (\ref{eq:con2}) indicates that $U_I$ is a global optimum solution to the problem in (\ref{eq:sobj}). As a result, Algorithm \ref{Alg:CMM} will converge to the global optimum of (\ref{eq:sobj}).
\end{proof}
\end{proposition}

\bibliographystyle{IEEEtran}
\bibliography{refMM}

\end{document}